\documentclass{sig-alternate}
\usepackage{multirow}
\usepackage{cite}
\usepackage{graphicx}
\usepackage[ruled,vlined]{algorithm2e}

\usepackage{amsthm}
\theoremstyle{definition}
\newtheorem{definition}{Definition}
\newcommand{\argmax}{\mathop{\rm argmax}\limits}

\begin{document}
\setlength{\pdfpagewidth}{8.5in}
\setlength{\pdfpageheight}{11in}

\setcopyright{acmcopyright}
\conferenceinfo{GECCO '15,}{July 11 - 15, 2015, Madrid, Spain}
\isbn{978-1-4503-3472-3/15/07}
\acmPrice{\$15.00}
\doi{http://dx.doi.org/10.1145/2739480.2754737}

\title{Optimization by Pairwise Linkage Detection, Incremental Linkage Set, and Restricted / Back Mixing:\\DSMGA-II}
\numberofauthors{2}
\author{
\alignauthor
Shih-Huan Hsu\\
       \affaddr{Taiwan Evolutionary Intelligence Laboratory}\\
       \affaddr{Department of Electrical Engineering}\\
       \affaddr{National Taiwan University}\\
       \email{r02921076@ntu.edu.tw}
\alignauthor
Tian-Li Yu\\
       \affaddr{Taiwan Evolutionary Intelligence Laboratory}\\
       \affaddr{Department of Electrical Engineering}\\
       \affaddr{National Taiwan University}\\
       \email{tianliyu@ntu.edu.tw}
       }

\maketitle

\abstract
This paper proposes a new evolutionary algorithm, called DSMGA-II,  to efficiently solve optimization problems via exploiting problem substructures. The proposed algorithm adopts pairwise linkage detection and stores the information in the form of dependency structure matrix (DSM). A new linkage model, called the incremental linkage set, is then constructed by using the DSM. Inspired by the idea of optimal mixing, the restricted mixing and the back mixing are proposed. The former aims at efficient exploration under certain constrains. The latter aims at exploitation by refining the DSM so as to reduce unnecessary evaluations. Experimental results show that DSMGA-II outperforms LT-GOMEA and hBOA in terms of number of function evaluations on the concatenated/folded/cyclic trap problems, NK-landscape problems with various degrees of overlapping, 2D Ising spin-glass problems, and MAX-SAT. The investigation of performance comparison with P3 is also included.


\section{INTRODUCTION}
Since the importance of linkage and problem decomposition has been addressed in the field of evolutionary computation \cite{Holland:1992:ANA:129194}, many algorithms that adopt linkage learning have been developed. In 2003, Yu \textit{et al.} \cite{Yu03agenetic} borrowed the concept of DSM from the organization theory and proposed DSMGA. One of its key mechanisms is pairwise linkage detection. The linkage information is then stored in a DSM and can be later used to construct different linkage models, such as the building-block (BB) graph \cite{Yu:2006:MAF:1293350,Yu:2006:CHD:1143997.1144210}. In 2010, Thierens \textit{et al.} proposed the optimal mixing operator (OM) and adopted it in the LT-GOMEA \cite{Thierens:2011:OME:2001576.2001661,Bosman:2012:LNO:2330163.2330247,Thierens:2010:LTG:1885031.1885060}. Unlike most traditional selection-crossover combination in GAs, the decision-making in OM is noise-free and hence results in a much smaller population-sizing requirement than most estimation-of-distribution algorithms. Combined with the linkage-tree model, the linkage tree genetic algorithm (LTGA) family has shown strong optimization ability on a wide range of problems \cite{Sadowski:2013:ULP:2463372.2463474,Thierens:2013:HPS:2463372.2463477,Luong:2014:MGO:2576768.2598261}. However, LT may not be the most proper linkage model for problem decomposition---recent researches involves LT pruning \cite{conf/gecco/BosmanT13, 6900275} and more expressive linkage models such as LN \cite{Bosman:2012:LNO:2330163.2330247}.
In this paper, we propose a new evolutionary algorithm. The proposed algorithm adopts the pairwise linkage detection from DSMGA and uses it to construct a new linkage model called the incremental linkage set (ILS). Inspired by the idea of OM, we propose two recombination operators, the restricted and the back mixing. Combining the DSM linkage information, ILS, and the two new mixing operators, the proposed algorithm empirically demonstrates stronger optimization ability than the LT-GOMEA and few other algorithms on the concatenated trap, folded trap, cyclic trap with overlapping, the NK-landscape problems with various degrees of overlapping, Ising spin-glass problems, and MAX-SAT.

The remainder of this paper is organized into three main parts. The first part revisits researches that are directly related to this paper. The second part details DSMGA-II and its operators. The third part describes the experiments and demonstrates the results, and finally conclusion follows. 

\section{RELATED WORKS}
In standard genetic algorithms, problems can be solved effectively and efficiently if solutions are mixed adequately. To match different kinds of problem structures, many techniques for building tunable models have been developed during the past decade. This section describes two important researches that we incorporated in our works.
\subsection{Dependency Structure Matrix Genetic Algorithm (DSMGA)}
Borrowed from the concept of organization theories, the dependency structure matrix genetic algorithm (DSMGA) utilizes DSM and clustering algorithms to detect interactions among variables. A DSM is essentially an adjacent matrix where each entry represents the pairwise information between two variables. 
The building-block information is then extracted by clustering the DSM. With the building-block information, DSMGA adopts building-block wise crossover instead of traditional gene-wise crossover. Such mechanisms have shown to be beneficial to the
search efficiency due to fewer disruptions of subsolutions.
\subsection{Optimal Mixing (OM)}
OM was first proposed as the recombination operator of the recombinative/gene-pool optimal mixing evolutionary algorithm \cite{Thierens:2011:OME:2001576.2001661}. OM applies function evaluation
during the recombination and take the change only if the fitness value improves. OM somewhat acts like building-block wise local search with interchangeable models. Models are usually described as the family of subsets (FOS) \cite{Thierens:2011:OME:2001576.2001661}. An FOS contains subsets of a certain set $S$. Each subset of FOS may appear more than once. The set $S$ contains all problem variable indexes, \textit{i.e.,} $\{1,2,\dots,{\ell}\}$. An FOS $\mathcal{F}$ can be written as an ordered set ${\langle}\textbf{\textit{F}}^1,\textbf{\textit{F}}^2,\dots,\textbf{\textit{F}}^\mathcal{|F|}{\rangle}$ where $\textit{\textbf{F}}^{\,i}{\subseteq}S$, i ${\in}\, \{1,2,\dots,\mathcal{|F|}\}$. Moreover, every problem variable index is contained in at least one subset in $\mathcal{F}$ to ensure that every linkage is used in mixing operators. OMEAs are superior to most GAs in many aspects. One of the noticeable properties is that a much smaller population size is required for optimization because of the noise-free decision-making in terms of the population sizing\cite{Goldberg91geneticalgorithms}. Furthermore, OM operators are capable of dealing with problems with overlapping structures efficiently with certain FOS\cite{ Bosman:2012:LNO:2330163.2330247}.

\section{DSMGA-II}
The following section gives details of the dependency structure matrix genetic algorithm II (DSMGA-II). Roughly speaking, DSMGA-II is the extension of DSMGA combined with the idea of OM and several new operators for improvement.  We first introduce its framework. The concept of DSM construction and incremental linkage set are described then, and finally two mixing operators are shown --- the restricted mixing and the back mixing. Note that all the algorithms in this paper are assumed to solve maximization problems for ease of expression.
\subsection{Framework of DSMGA-II}
The major components of DSMGA-II are linkage information retrieval via pairwise detection, expressive linkage model construction, and efficient mixing that balance between exploration and exploitation. The idea of pairwise linkage detection can be backtracked as early as LINC and LIMD ~\cite{munetomo1999identifying}, and is then adopted during model building in DSMGA. Original LTGA does not adopt pairwise linkage detection until a later work~\cite{pelikan2011pairwise} which shows the pairwise approximation outperforms the original. Similar to DSMGA, DSMGA-II  adopts pairwise detection due to its resistance of sampling noise.  The algorithm first utilizes pairwise linkage detection and stores the information in DSM. After the building-block information is obtained, the mixing operators then proceed with incremental linkage set, which can be seen as a specific type of FOS. In addition, since local search has shown to be beneficial to EDAs\cite{Chen:2013:EDH:2463372.2463418}, a bit-flipping local search operator is therefore performed right after the initialization of population to improve the quality of model building. The DSM is updated at the beginning of each generation with the after-selection population. The selection is tournament selection with the selection pressure $2$ according to the suggested results in \cite{Yu:2007:PSE:1276958.1277080}. To prevent from overfitting due to frequent model building, the algorithm updates DSM once every $O(\ell)$ generations. Note that the population after-selection is only used for updating the DSM; the rest of algorithm proceeds with the original population.

The pseudo-code is given in Algorithm~\ref{dsm2}. Each operator is detailed in the following sections. The population is
denoted by ${\mathcal{P}}$, with the chromosomes $P_1,\dots , P_{|\mathcal{P}|}$, the problem size is $\ell$, and the after-selection population is ${\mathcal{S}}$. $R$ is a constant proportional to $\ell$. The
DSM is denoted by ${\mathcal{D}}$. ${\mathcal{I}} = {\langle}{I_1},{I_2},\dots{I_{\mathcal{|P|}}{\rangle}}$ is a random permutation of $\{1,2,\dots{\mathcal{|P|}}\}$, and the incremental linkage set is denoted by ${\mathcal{L}}$, which is a set of masks and is elaborated in the following section. Each iteration of the ``while" loop corresponds to a generation, and the algorithm terminates when the optimal solution appears.

\begin{algorithm}[tb]
\label{dsm2}
 ${\mathcal{P}}$: population, $\ell$: problem size ${\mathcal{D}}$: DSM, ${\mathcal{L}}$: incremental linkage set, $R$: constant\\
 \BlankLine
 randomly initialize population ${\mathcal{P}}$\\
 ${\mathcal{P}} \gets$ \textsc{RunLocalSearch(${\mathcal{P}}$)}\\
 \While{$\neg$\textsc{ShouldTerminate}}{

  ${\mathcal{S}} \gets$ \textsc{TournamentSelection(${\mathcal{P}}$,  $s$)}\\
  ${\mathcal{D}} \gets$  \textsc{UpdateMatrix(${\mathcal{S}}$)}\\
   \For{$k=1$ to $R$}{
  ${\mathcal{I}}\gets$ random permutation from 1 to |${\mathcal{P}}$|\\

  \For{$i=1$ to $|{\mathcal{P}}|$}
  {
   $(P_{I_i},L_s) \gets$ \textsc{RestrictedMixing($P_{I_i}$)}\\
     ${\mathcal{P}} \gets$ \textsc{BackMixing($P_{I_i},L_s$)}\\
  }

}
 }
 \Return the best instance in $\mathcal{P}$\\
 \caption{DSMGA-II}
\end{algorithm}

\subsection{Linkage Learning Model: ILS}

DSM is a graph representation of the dependency between two variables, where each entry $e_{ij}$ is the measure of dependency between nodes $i$ and $j$. Entries can be real numbers or integers. The greater the $e_{ij}$ is, the more significant measure between nodes $i$ and $j$ is. Once the matrix is constructed, the information can be interpreted as a graph, where vertices are the variables and edges are the measures of dependency.

\theoremstyle{definition}
\begin{definition}
(\textit{Maximum-weight connected subgraph}, \textbf{MWCS}) \textit{Given an undirected graph $G = (V,E)$ with edge weights $w:E{\rightarrow}\,\mathbb{R}$. MWCS is a complete subgraph $G^{\prime}=(V^{\prime},E^{\prime})$ of $G$, $(i.e.,\  V^{\prime}{\subseteq}{V}$, $\ E^{\prime}{\subseteq}{E},\ {\forall}\,u,v\,{\in}\,V^{\prime}{\Rightarrow}\,(u,v)\,{\in}\,E^{\prime})$ such that $\sum_{e{\in}E^{\prime}}w(e)$ is maximum.}
\end{definition}

	In this paper, pairwise dependency is measured by the mutual information\cite{kullback1951}, which can be calculated as follows:
\begin{equation}
I(X;Y) = \sum_{y \in Y} \sum_{x \in X} p(x,y) \log{ \frac{p(x,y)}{p(x)\,p(y)}}\ ,                              \end{equation}
where $X$ and $Y$ are two random variables, and $x$ and $y$ are outcomes of the random variables respectively. $p(x,y) = p(x)p(y)$ if $X$ and $Y$ are independent, and thus $I(X;Y)$ equals to 0.

Many clustering algorithms for DSM have been proposed. LTGA~\cite{Thierens:2010:LTG:1885031.1885060} applies a hierarchical clustering technique with normalized variation of information to build the linkage tree model of population. The clustering technique was further simplified by using pairwise linkage information with average linkage clustering to achieve efficient computation without apparent drawbacks\cite{Thierens:2011:OME:2001576.2001661}.\par
We simply use mutual information as the dependency measure, which is the same as that of DSMGA. However, instead of clustering variables, we look for a specific subgraph called approximated maximum-weight connected subgraph (AMWCS) to construct the model. The concept of AMWCS is similar to MWCS\cite{NAV:NAV4, mwcs}, but AMWCS is constructed by greedy approach. The algorithm initializes an AMWCS from a certain vertex and iteratively adds one vertex that is most related to the current AMWCS into AMWCS with average mutual information. The vertex for insertion in each iteration is chosen by the following equation:
\begin{align}
j=\argmax_{j{\in}C^\prime}{\frac{1}{|C|}} \sum_{k{\in}C} I(j,k)\ ,
\end{align}
where $C$ is the set of vertices in the current AMWCS and $C^\prime$ is the set of all vertices but those in $C$.\par

\theoremstyle{definition}
\begin{definition}
(\textit{Incremental linkage set}, \textbf{ILS})\\ \textit{ ILS is an FOS  $\mathcal{F}={\langle}\textbf{\textit{F}}^{\,1},\textbf{\textit{F}}^{\,2},\dots,\textbf{\textit{F}}^\mathcal{\,|F|}{\rangle}$ that satisfies\\ $\forall \,i,j \in \{1, 2,\dots, \mathcal{|F|}\}$, $i < j\ {\Rightarrow}\ \textbf{\textit{F}}^{\,i}\subset \textbf{\textit{F}}^{\,j}$.}
\end{definition}

The incremental linkage set is a specific type of FOS which consists of the AMWCSs after every iteration of insertion, thus the size of each element sequentially increases by 1 in this case. According to the results in \cite{Yu:2006:MAF:1293350}, one of the keys to achieve efficient recombination on problems with overlapping structures is that every pair of building-blocks should have a substantially high enough probability to be mixed. Therefore, a vertex is first randomly chosen as the initial AMWCS. For instance, consider a problem with length 5, and the vertex 3 was randomly chosen from $\{1,2,3,4,5\}$ at first (Figure 1). After 4 iterations, the vertices were inserted into AMWCS in the order of 3--1--5--4--2, and the incremental linkage set $\mathcal{L}={\langle}\{3\},\{3,1\},\{3,1,5\},\{3,1,5,4\},\{3,1,5,4,2\}{\rangle}$.
Note that ILS is an ordered set.\\

\begin{figure}[h]
\label{fig:mw}
\centering\includegraphics[width=3in]{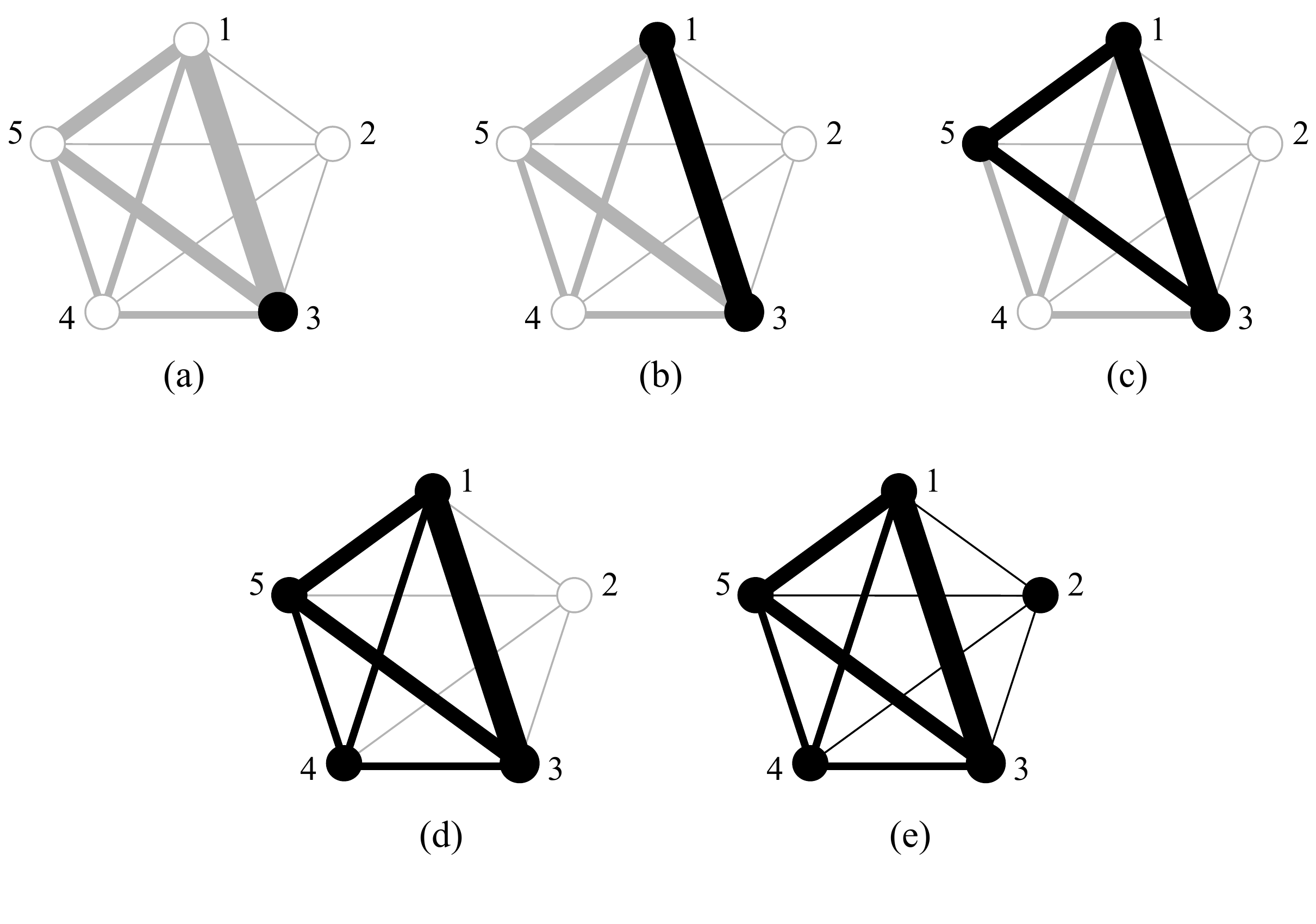} 
\caption{(a) to (e) shows the AMWCS in each iteration. The vertices represent variables and the gray edges represent the dependency measure between pairs. The larger the measure is, the wider the edge is. Black vertices and edges represent the determined AMWCS.}
\end{figure} 

\subsection{Restricted and Back Mixing}

$\mbox{DSMGA-II}$ adopts two mixing operators: the restricted mixing and the back mixing.  Unlike canonical genetic algorithms (GAs), $\mbox{DSMGA-II}$ does not actually generate offspring by recombination of parent solutions. Instead, it applies the restricted mixing to flips bits with the masks. Each mask is essentially a set of indexes that indicates which variables should be considered together during mixing operations. Relatively large mutual information implies that the corresponding bits are different from that of other solutions in current population. In other words, the bit will not be flipped if it converges to the same value. Also, when two solutions exchange subsolutions with certain mask, each allele may not actually vary because of the same subsolutions, especially for the population that is getting on for convergence. Although this drawback may not increase the number of function evaluations (NFE), it is still time-consuming. The flipping process however ensures that every trial is different. To sum up, the restricted mixing starts from choosing a receiver for the trial solution. Subsolutions in the trial solution are flipped with masks in ILS from small size to large size, and the change is preserved if the fitness does not decrease as well as the trial solution is unique in population. Once the solution improves, the restricted mixing terminates.\par
Moreover, in terms of building-block supply\cite{Goldberg91geneticalgorithms}, we believe that proper subsolutions exist in current population. Accordingly, the restricted mixing also terminates when the complementary pattern of receiver does not exist in the current population. This can also be seen as a mixing between receiver and the chromosome which contains the complementary pattern of receiver, and this is the reason for calling this operation
restricted mixing. The pseudo-code for the restricted mixing is given in Algorithm~\ref{rm}. The population is
denoted by ${\mathcal{P}}$ with problem size $\ell$. The incremental linkage set ${\mathcal{L}}$ = ${\langle}L_1,L_2,\dots,L_{\mathcal{|L|}}{\rangle}$. $P_L$ is the pattern of chromosome $P$ selected with mask $L$. $P_L^{\prime}$ is the complementary pattern of $P_L$. $T$ is the trial solution, and the evaluation function is $f$. $C$ is the set of vertices in AMWCS.

\begin{algorithm}
\label{rm}
${\mathcal{P}}$: population, $\ell$: problem size, \\$C$: set of vertices in AMWCS\\ ${\mathcal{L}}$: incremental linkage set, $f$: evaluation function,\\ $T$: trial solution\\
\KwIn{$P:$ receiver}
\BlankLine
 AMWCS $\gets$ random number from 1 to $\ell$\\
 \While{$P_C^{\prime}\ \in\ \mathcal{P}$  }{$\mathcal{L} \gets \mathcal{L} \cup C$\\join the nearest vertex into AMWCS\\
 }
  \For{$i=1$ to $|\mathcal{L}|$}
  {
   $T \gets P$\\
   $T_{L_i} \gets T_{L_i}^{\prime}$\\
     \If{$f(T){\geq}f(P)\ and\ {T} \notin \mathcal{P}$}{$P \gets T$\\
     \Return $(P,L_i)$}
  }

 \caption{Restricted Mixing}
\end{algorithm}

 Figure 2 is an example of the restricted mixing. Consider an ILS, a population $\mathcal{P}=\{P_1,P_2,\dots,P_5\}$, and a receiver $P=P_4$.

\begin{figure}[h!] 
\centering\includegraphics[width=3.3in]{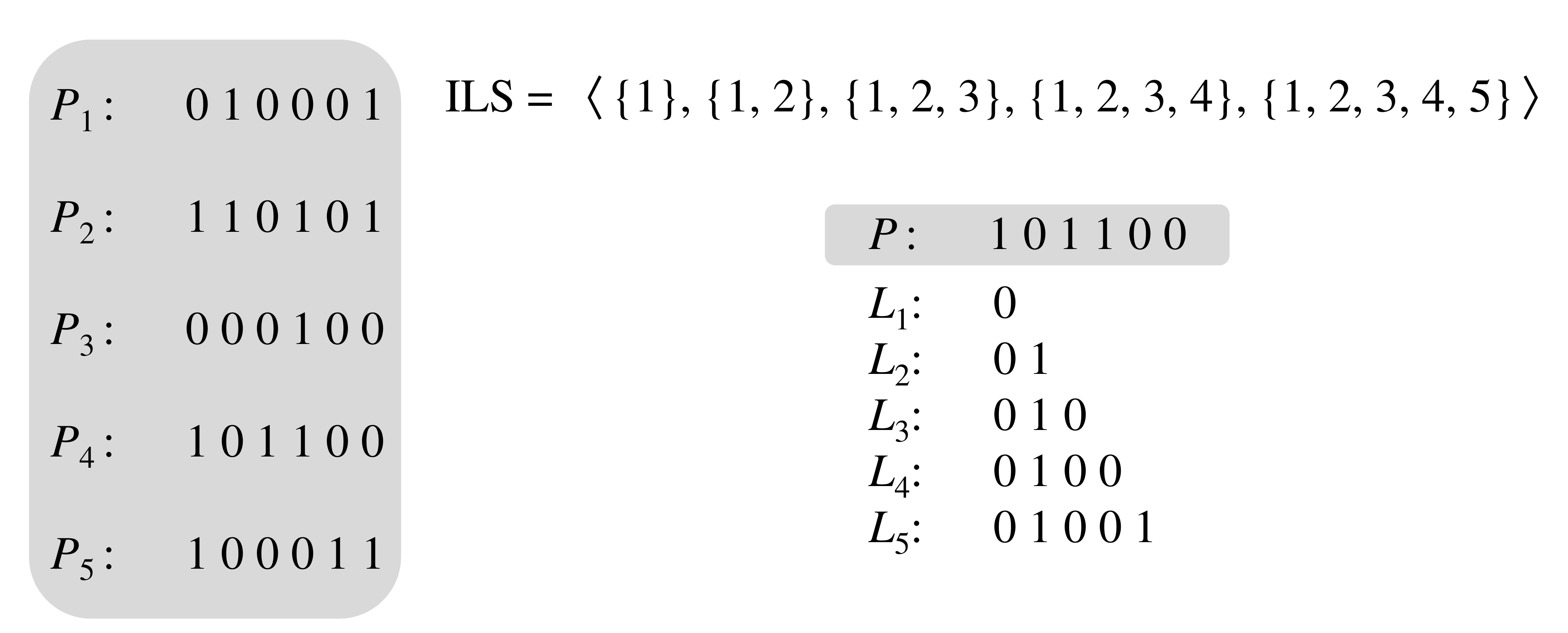} 
\caption{An example of the restricted mixing. Population $\mathcal{P}$ is on the left, and receiver $P=P_4$ is on the right. The complementary pattern with the first mask $\{1\}$ of receiver is 0,  which exists in chromosome 1 and 3. With the second mask $\{1,2\}$, the complementary pattern is 01, which exists in chromosome 1 only, and so on. However, the complementary pattern with the fifth mask $\{1, 2, 3, 4, 5\}$ is 01001, which does not exist in any chromosome of the population. Therefore, the last mask that should be utilized is the fourth one.} 
\end{figure}

%

\begin{algorithm}
\label{bm}
${\mathcal{P}}$: population, $f$: evaluation function, $T$: trial solution\\$E$: set of candidate solutions\\
  \KwIn{$D:$ donor,\ $L$: mask}
  \BlankLine
  $improved \gets false$\\
  \For{$j=1$ to $|\mathcal{P}|$}
  {
   $T \gets P_j$\\
   $T_L \gets D_L$\\
     \If{{$f(T)>f(P_j)$}}
     {$P_j \gets T$\\
     $improved \gets true$}
     \Else{\If{{$f(T)=f(P_j)$}}
     {$E \gets E \cup \{T\}$}}
  } 
 \If{$\neg improved$}
     {$accept\ all\ solutions\ in\ E$}
 \Return $\mathcal{P}$\\
 \caption{Back Mixing}
\end{algorithm}

After the restricted mixing finishes, the successfully flipped receiver during the restricted mixing becomes the donor of the back mixing. Every chromosome in the population is then mixed with the flipped pattern, and the change is adopted only if the fitness is improved. Note that the acceptance criterion is different from the restricted mixing. Real-world problems may contain various landscapes such as plateaus and basins, where solutions are of equal qualities and difficult to explore. Many operators have been developed to deal with such difficulties, such as the forced improvements (FI)\cite{Bosman:2012:LNO:2330163.2330247}. However, diversity issue should be handled carefully. The back mixing tempts to substitute every solution with the same allele fragment, and it causes a strong drift effect if side walks (changes taken for equal fitness) are allowed. In contrast, if the acceptance criterion is set to strict fitness improvement, more evaluations are needed in order to jump out of the plateaus. Either case decreases the performance. Therefore, the acceptance criterion for the back mixing is set to strict fitness improvement as default, and side walks are allowed only if no solution is improved with the default setting during the back mixing. The pseudo-code for the back mixing is given in Algorithm~\ref{bm}. Note that the implementation is slightly different from the above to alleviate the computational burden. In short, the idea behind this operator is graph refining to reduce unnecessary evaluations. The empirical results suggest that the back mixing is able to deal with both plateaus and diversity issues, which is detailed in Section 4.3. 
\section{TEST PROBLEMS AND EXPERIMENTS}
This section first describes the benchmark problems used in this paper. The setup of the experiments is then given, and followed by experimental results and discussions.
\subsection{Optimization Problems}
In our research, six types of linkage benchmark problems are considered, including four classic linkage-underlying problems and two real-world problems. In the following context, the number of variables in the test function is referred to as the problem size, denoted by $\ell$, and a chromosome is denoted as a vector $\mathbf x = \left(\mathbf x_1 ,\mathbf x_2,\dots,\mathbf x_\ell \right) $.
\subsubsection*{Concatenated trap}
The concatenated trap is composed of $m$ additively separable trap functions, and each of which contains $k$ variables~\cite{mktrap}. The number of variables of the concatenated trap function is $m\cdot k$. It is well-known that the problem can only be efficiently solved when the underlying structure is detected and preserved while mixing \cite{thierens1993mixing}.
Its fitness is as follows:
\begin{equation}
f^{trap}_{m,k} \left({\mathbf x}\right)
= \sum_{i=1}^{m} f^{trap}_{k} \left( \sum_{j=i\cdot k - k + 1}^{i \cdot k} {\mathbf x}_j \right),
\end{equation}
where
\begin{equation}
f^{trap}_{k} \left(u\right) 
= \left\{
\begin{array}{l l}
1 & \text{if $u=k$}, \\
\frac{k-1-u}{k} & \text{otherwise}.
\end{array}
\right.
\end{equation}
\subsubsection*{Cyclic trap}
\label{subsec:ctrap}
The cyclic trap is composed of overlapping trap functions with wraparound~\cite{Yu:2005:LLO:1068009.1068209}.
The fitness function is given as follows:
\begin{equation}
f^{cyclic}_{m,k} \left({\mathbf x}\right)
= \sum_{i=1}^{m} f^{trap}_{k} \left( \sum_{j=i\cdot \left(k-1\right) - k + 2}^{i \cdot \left(k-1\right)+1} {\mathbf x}_j \right),
\end{equation}
where
\begin{equation}
f^{trap}_{k} \left(u\right) 
= \left\{
\begin{array}{l l}
1 & \text{if $u=k$}, \\
\frac{k-1-u}{k} & \text{otherwise}.
\end{array}
\right.
\end{equation}
and
\begin{equation}
{\mathbf x}_j = {\mathbf x}_{j-\ell},\ \textrm{if}\ \ell<j\leq 2\ell.
\end{equation}
In the cyclic trap, the linkage information needs to be carefully taken care of. Take a 12-bit cyclic trap with $k=5$ as an example. Although the size of the subfunction is 5, the fitness of 111110000000 (one correct subsolution, $f=2.2$) is lower than that of 000000000000 (all incorrect subsolutions, $f=2.4$). One may of course increases the sizes of masks during recombination; however, the risk of doing so is the increase of the number of trials and hence NFE increases.
\subsubsection*{Folded trap}
There are many variants of folded trap~\cite{Goldberg92massivemultimodality}, and we use the bipolar deceptive function with $k=6$ in our experiment. The folded trap contains two global optima and many local optima. The key difficulty of this problem is that local optima reside on plateaus, and hence the exploration over plateaus needs to be performed. However, unnecessary exploration, such as from $11111*\ldots*$ to $00000*\ldots*$, increases NFE. Additionally, the folded trap is symmetric. Any preference toward 0s or 1s does not help find the global optimum. The fitness function with $k=6$ is described as follows:

\begin{equation}
f^{folded}_{m,k=6} \left({\mathbf x}\right)
= \sum_{i=1}^{m} f^{folded}_{k=6}\left( \sum_{j=i\cdot k - k + 1}^{i \cdot k} {\mathbf x}_j \right) ,
\end{equation}
where
\begin{equation}
f^{folded}_{k=6} \left(u\right) 
= \left\{
\begin{array}{l l}
1 & \text{if $ |u-3|=3$}, \\
0.8 & \text{if $|u-3|=0$}, \\
0.4 & \text{if $|u-3|=1$}, \\
0 & \text{if $|u-3|=2$}. \\
\end{array}
\right.
\end{equation}


\subsubsection*{NK-landscape}
The NK-landscape functions are composed of overlapped, randomly generated subfunctions~\cite{Pelikan:2009:PEA:1569901.1570018}.
There are three parameters of the function: $\ell$, $k$, and $s$, where $\ell$ is the problem size, $k$ is the number of neighbors of one gene, and $s$ is the step size, the offset of two adjacent subfunctions.
The function of the NK-landscape is given as follows:
\begin{equation}
f^{NK}_{\ell,k,s}\left(\mathbf x\right)
= \sum_{i=0}^{(\ell-k-1)/{s}} f^{subNK}_{k,i} \left( {\mathbf x}_{i\cdot s+1}, {\mathbf x}_{i\cdot s+2}, \dots {\mathbf x}_{i\cdot s+k+1} \right),
\end{equation}
where $f^{subNK}_{k,i}$ are subfunctions subject to the constraint that $f^{subNK}_{k,i} \left({\mathbf x}\right) \in [0,1]$ for any valid input $\mathbf x$.
The NK-landscape functions are commonly considered as general cases of problems due to its random problem landscapes. One may change the degree of overlapping of NK-landscape by varying the step size $s$.

\subsubsection*{Ising spin-glass}
The Ising spin-glass is a well known problem of statistical mechanics. Given a set of variables that represent spins, each one is in one of two states of $\{+1,-1\}$. For any two adjacent spins $i$ and $j$, there is an coupling constant $J_{ij}$. The objective is to find a state of spins called $ground\ state$ for given constants $J_{ij}$ that minimizes the energy of system. The energy (fitness) function is given as follows:
\begin{equation}
f^{spin}_n \left(\mathbf x\right)
= -\sum_{i,j=0}^{n}{\mathbf x}_i{\mathbf x}_j J_{ij}\ .
\end{equation}
We consider a special case of spin-glass in our experiments: The spins are arranged on 2-D grid with each spin interacts with only four nearest neighbors on the grid, and the coupling constants contain only two values, $J_{ij} \in \{+1,-1\}$. Ising spin-glass systems are usually studied due to their particular properties, such as symmetry (\textit{i.e.}, the fitness remains unchanged when swapping 0 and 1) and several plateaus.
\subsubsection*{MAX-SAT}
The maximum satisfiability problem (MAX-SAT) was the first problem proven to be NP-complete. The problem consists of a series of $logical\ and$ clauses, where each clause is a series of $logical\ or$ variables. Each variable represents either a predicate or a negation of a predicate. The MAX-SAT problem can be described as following conjunctive normal form (CNF) formula:
\begin{equation}
F=\bigwedge_{i=1}^m \left(\bigvee_{j=1}^{k_i}{l_{ij}}\right) ,
\end{equation}
where $m$ is the number of clauses, $k_i$ is the number of literals in the $i$-th clause, $l_{ij}$ is the $j$-th literal in the $i$-th clause, which corresponds to a gene in the chromosome. The fitness of $\mathbf x$ is the number of clauses in $F$ that are satisfied under the interpretation $\mathbf x$. For our experiments, we use the Uniform Random-3-SAT instances from SATLIB\footnote{http://www.cs.ubc.ca/$\sim$hoos/SATLIB/benchm.html} with all satisfiable clauses.
\subsection{Experiment Setup}
Since $\mbox{DSMGA-II}$ incorporates the concepts from DSMGA and OM, it is intuitive to compare $\mbox{DSMGA-II}$ with them. However, original DSMGA~\cite{Yu03agenetic} does not deal with problems with overlapping structures. Even combined with other techniques~\cite{Yu:2005:LLO:1068009.1068209, Yu:2006:MAF:1293350}, DSMGA still cannot rival \mbox{DSMGA-II} and hence is left out in this paper. 
Concerning GOMEAs, several different linkage models have been proposed~\cite{conf/gecco/BosmanT13, Bosman:2012:LNO:2330163.2330247} since the original one, but the improvements are merely marginal, and LT-GOMEA is still considered as state-of-the-art. 

Furthermore, we also compare DSMGA-II with the hierarchical Bayesian optimization algorithm (hBOA)~\cite{Pelikan03hierarchicalboa} since it is a milestone and is often compared with in recent EDA researches. Finally, the parameter-less population pyramid (P3)~\cite{Goldman:2014:PPP:2576768.2598350} has shown outstanding performance and worths comparing with. However, P3 is extraordinary from the traditional EA framework, which makes designing a fair comparison virtually impossible. Hence we leave the comparison to the end of this section.

Traditionally, most researches utilize a bisection procedure \cite{pelikan2005hierarchical} to find the minimum population size that is sufficiently large for a certain number of consecutive successful convergences; the required NFE at that population size is then used as the performance measure. However more than often, such minimum population sizing does not yield the minimum NFE as desired, and the differences vary for different algorithms. 

For fair comparisons, we adopt an adaptive sweeping procedure to find the minimum NFE. The procedure starts by sweeping the population size through a reasonable range with a predefined step. The average NFE is then recorded over a certain number of consecutive successful hits. If the algorithm fails to converge to the global optimum with the population size, NFE is recorded as infinity. The sweeping range is then narrowed around the population size that yields the minimum NFE with a smaller step. The procedure iterates until the sweeping range becomes small enough. In this paper, the requirement is 10 consecutive successful hits, the initial population size is 10, the initial step size is 30 and then is divided by 2 for each iteration, and procedure terminates when the sweeping range is within 5\% of the population size. For each problem, the results are averaged over 100 independent runs. 

The experimental settings of algorithms are follows. The selection pressures of DSMGA-II, LT-GOMEA, and hBOA are set to 2. In DSMGA-II, the model building is performed once every $\ell/50$ generations. In this paper, LT-GOMEA stands for the version with forced improvements~\cite{conf/gecco/BosmanT13}, implemented by the inventors of GOMEA\footnote{http://homepages.cwi.nl/$\sim$bosman/source\_code.php}. Also, LT-GOMEA is performed without local searcher, because its efficiency evidentially decreases by doing so~\cite{Bosman:2011:RLS:2001858.2002065}. 

For the NK-landscape problem, we choose NK-S1 (the highest degree of overlapping), S3 and S5 (non-overlapping) (\textit{i.e.}, parameter $s$ = 1, 3, 5) with 100 randomly generated instances each, and parameter $n$ is set to 4. In this way, we can check how degrees of overlapping affect the performances of the algorithms. For the concatenated trap and the cyclic trap, the subfunction size $k$ is set to 5, and the $k$ of folded trap is set to 6.

\subsection{Results and Discussions}

On all problems that we considered, $\mbox{DSMGA-II}$ requires the least NFE than that of others. In the following sections, the results of traditional EAs are detailed at first, and the comparison with P3 is then given.  


As shown in Figure 3, the results indicate that the differences between DSMGA-II and others become larger as the degree of overlapping decreases. The back mixing spends a few function evaluations on refining the graph at first. As the linkage information becomes clearer, the correct subsolution stands out, and the back mixing makes it quickly dominate the population. This makes the algorithm more efficient. However, if the problem structures are severely overlapped, the graph refining process is prolonged and thus NFE increases. 

\begin{figure}[h!] 
\centering\includegraphics[width=3.4in]{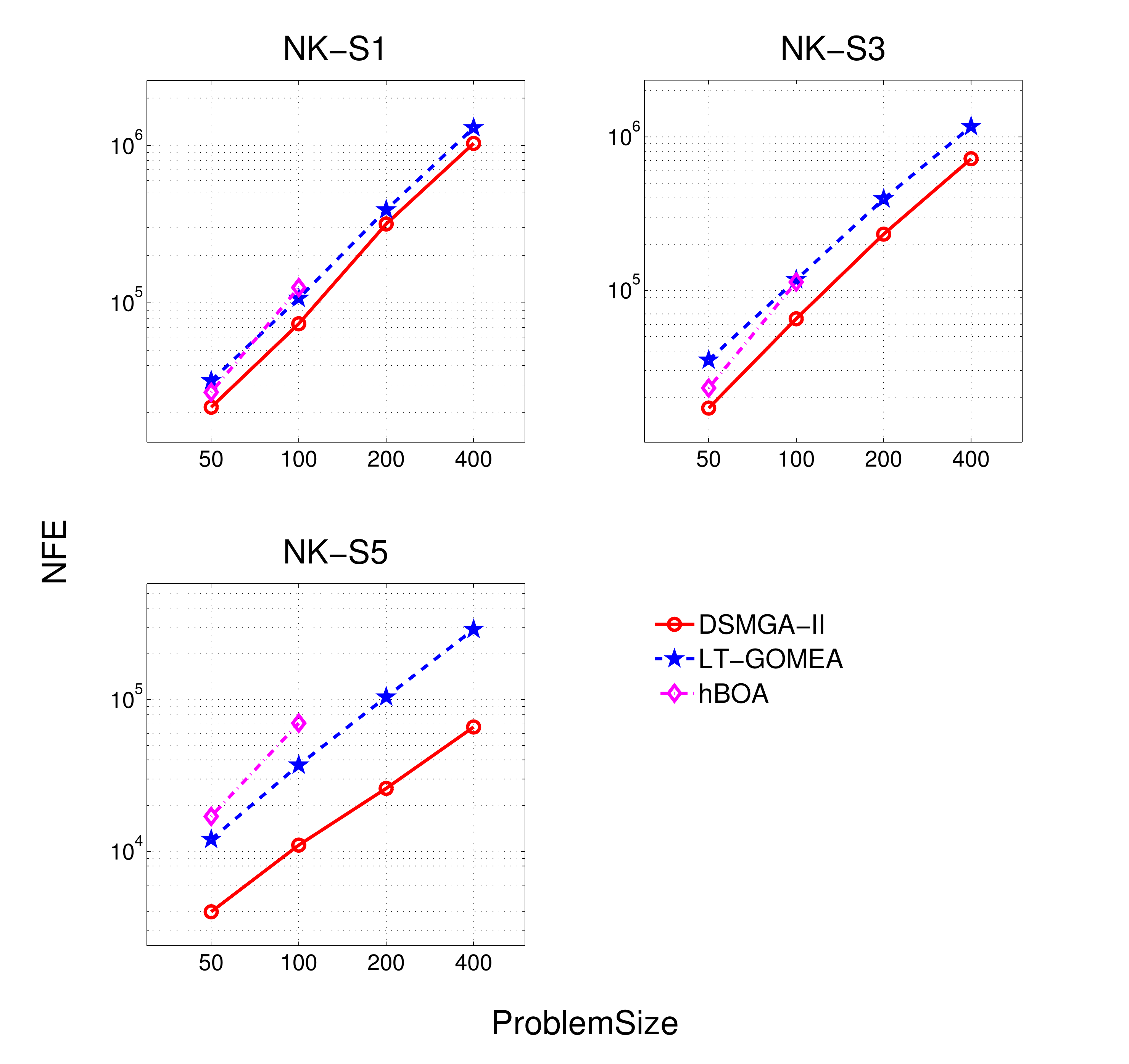} 
\caption{Scalability of DSMGA-II, LT-GOMEA and hBOA on NK-landscape problems with various degrees of overlapping.} 
\end{figure}

The results also demonstrate that $\mbox{DSMGA-II}$ is capable of handling problems with overlapping structures, even for the randomly generated problem landscapes. This is indicative that ILS model expresses overlapping relations well. A possible scenario of model building is shown in the Figure 4.

\begin{figure}[h] 
\centering\includegraphics[trim=50mm 40mm 40mm 0,width=2.3in]{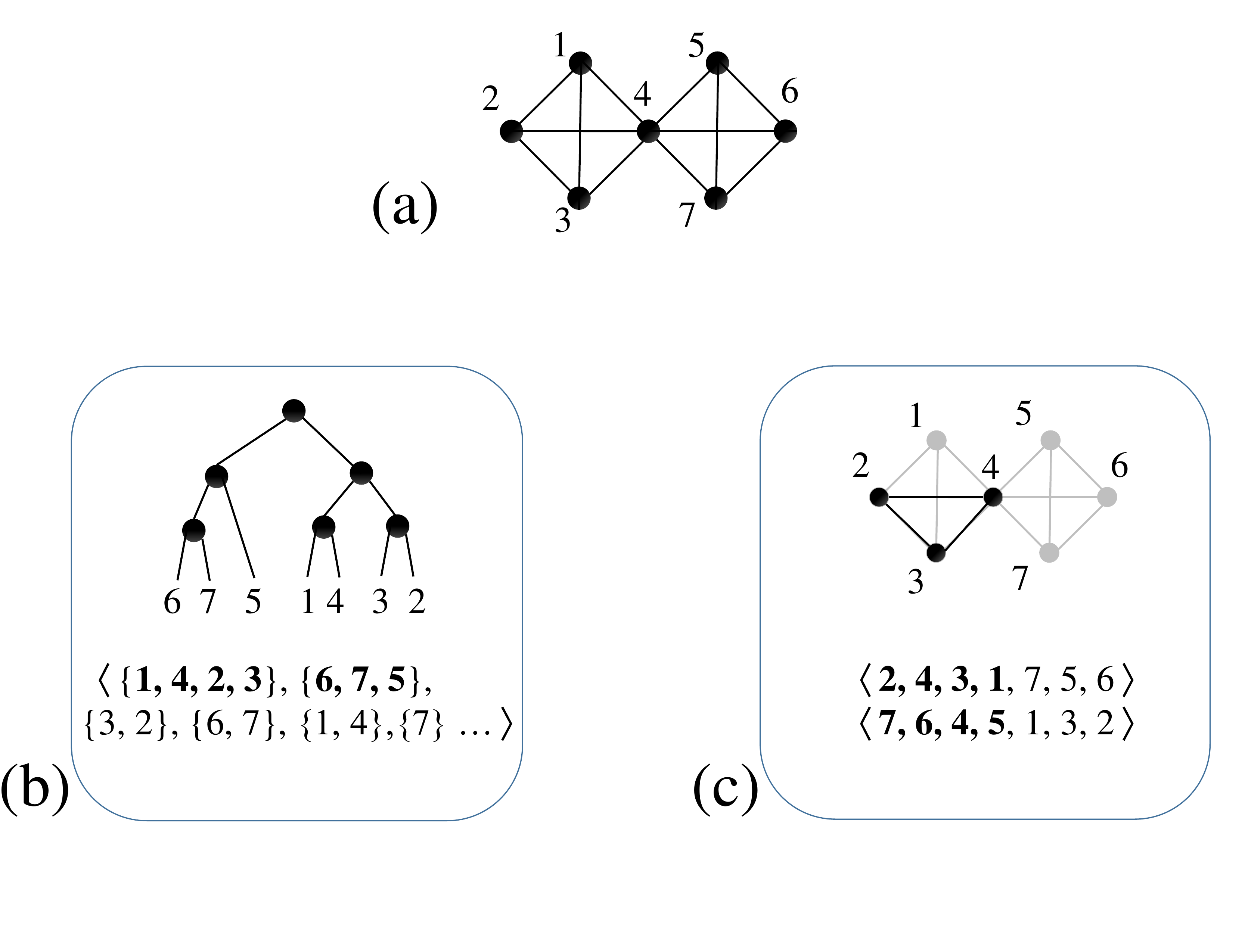} 
\caption{A snapshot of model building. (a) is the real structure of problem with vertex 4 overlapped. (b) is a LT model constructed from (a), and (c) is an ILS model with some possible AMWCS constructing sequences. Both LT and ILS model capture the problem structure. However, mask \{4, 5, 6, 7\} occurs in ILS within a single generation, while LT can only obtains this mask in later generations by rebuilding the model.} 
\end{figure}


The results of various types of deceptive problems are shown in Figure 5. For the concatenated trap, the slope of $\mbox{DSMGA-II}$ decreases as the problem gets larger. 

For the cyclic trap, the scalability of all algorithms appears very similar, with $\mbox{DSMGA-II}$ being lower by a constant factor. As mentioned in Section~\ref{subsec:ctrap}, the cyclic trap cannot be solved efficiently merely with the correct problem decomposition. The usage of linkage information is also the key. ILS automatically extends the mask for trials and stops on first successful recombination to avoid spending unnecessary function evaluations, while many other EAs utilize the linkage models determined by certain thresholds, which are sensitive to the parameter settings.

The folded trap contains a large number of local optima that reside on plateaus. Performing an efficient search without losing too much diversity is the key to conquer such difficulty. The back mixing leads to drift of subsolutions as necessary, and the restricted mixing checks if the trial solution is unique in population to keep diversity. $\mbox{DSMGA-II}$ as a result shows the good ability on dealing with attractions without trading efficiency for diversity.

\begin{figure}[!h] 
\centering\includegraphics[width=3.4in]{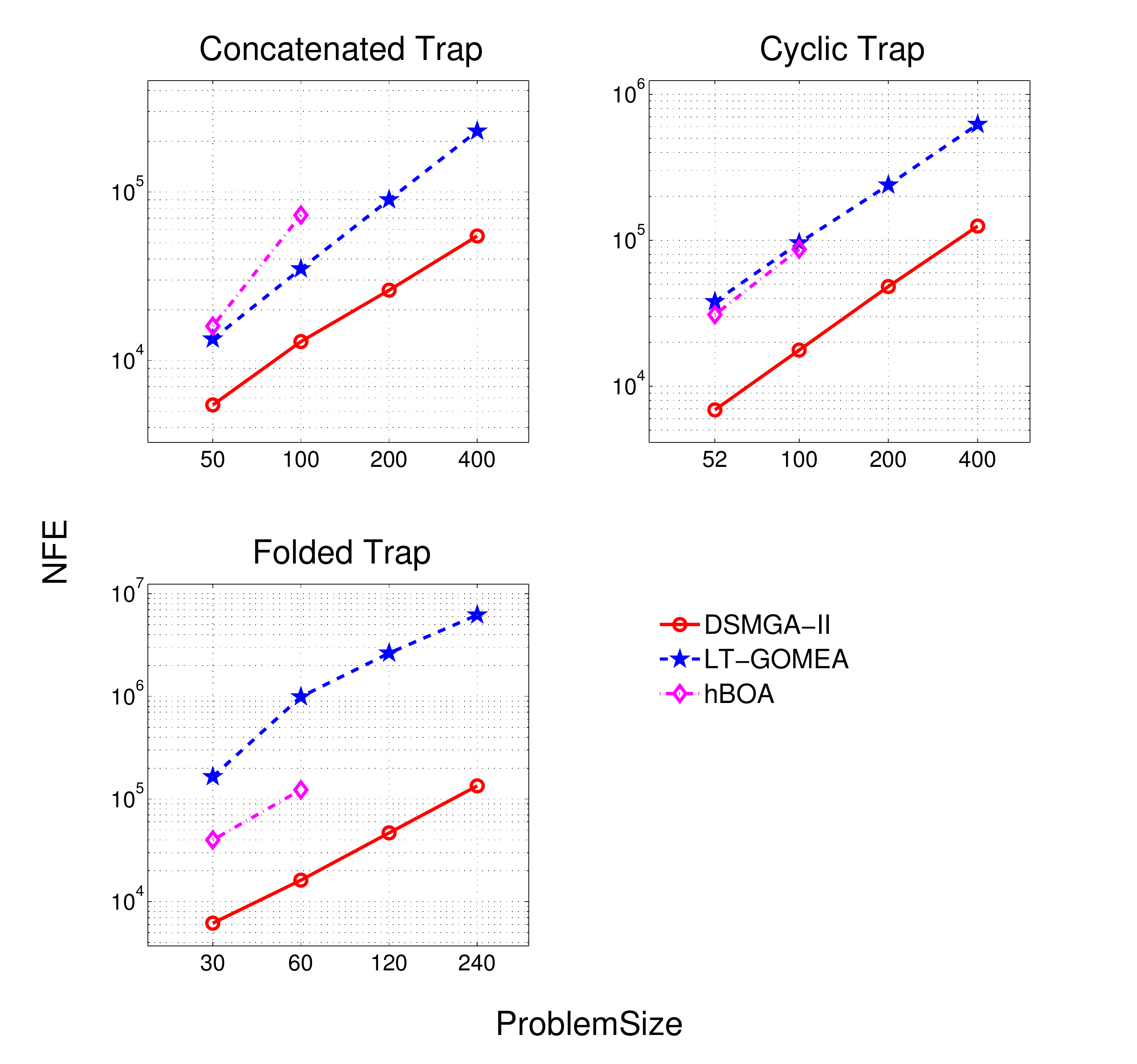} 
\caption{Scalability of DSMGA-II, LT-GOMEA and hBOA on the problems of deceptive variants.} 
\end{figure}


For Ising spin-glass problems, the slope of $\mbox{DSMGA-II}$ decreases as the problem gets larger (Figure 6). This is probably because the problems contain a large number of plateaus, where the advantages of $\mbox{DSMGA-II}$ hold. The overall time spent in fitness evaluations appears to grow polynomially as $O(n^3)$  for $\mbox{DSMGA-II}$, which is close to the best known results of problem-specific algorithm for Ising sping-glass~\cite{151468,galluccio1999theory}.
For MAX-SAT problems, NFE of $\mbox{DSMGA-II}$ still grows exponentially although it requires the fewest function evaluations.

\begin{figure}[!h] 
\centering\includegraphics[trim=0 110mm 0 0, width=3.4in]{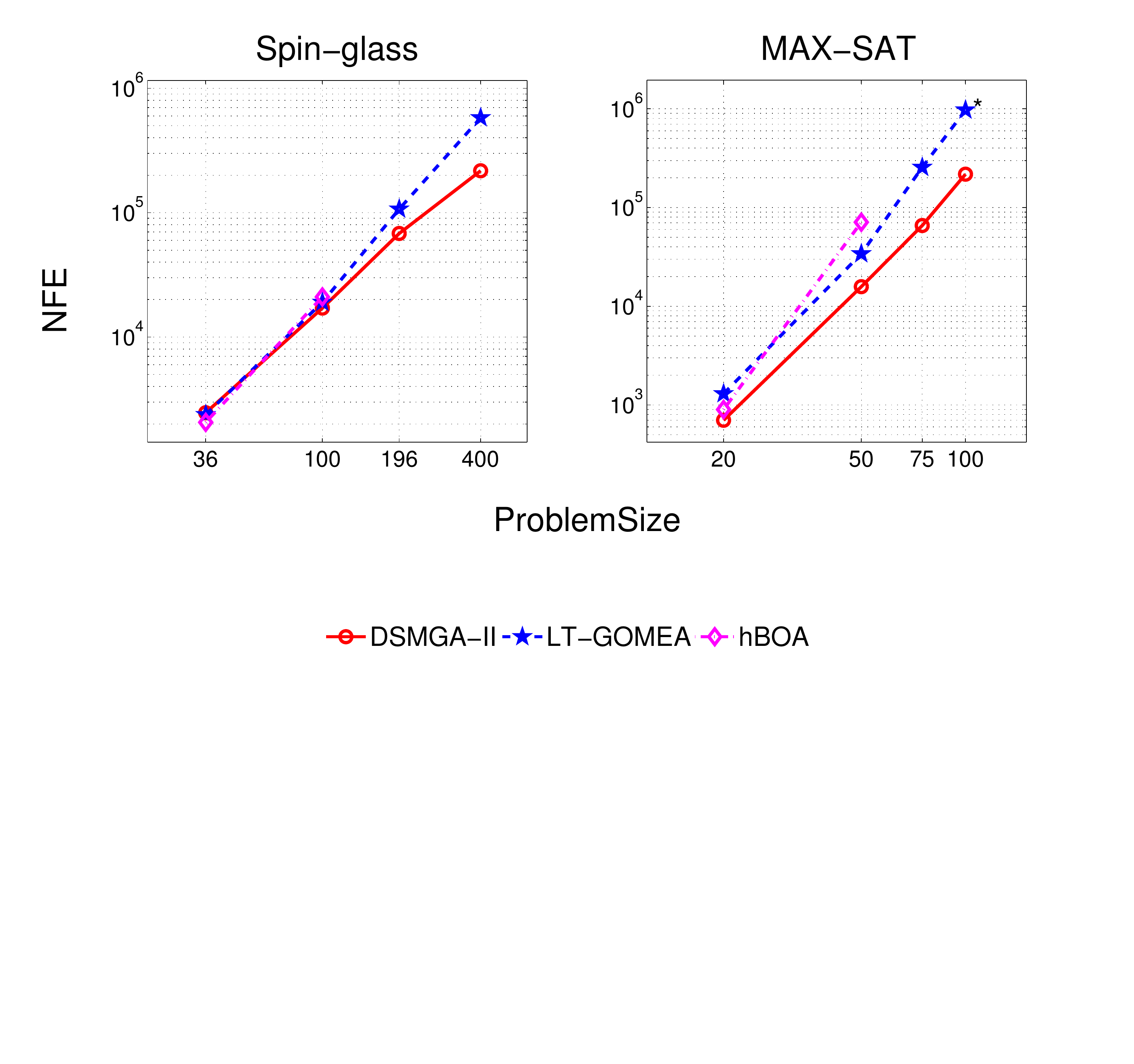} 
\caption{Scalability of DSMGA-II, LT-GOMEA and hBOA on Spin-glass and MAX-SAT (*LT-GOMEA fails to reach the global optima for two instances with $\ell=100$ on MAX-SAT).} 
\end{figure}

As mentioned before, P3 differs from traditional EAs, and a fair comparison is difficult. P3 does not need a predetermined population size; instead, it increases the population one-by-one when needed until it hits the global optimum. In traditional EA framework, the above situation is similar to a sweeping procedure with step size of 1, and the convergence requirement is merely 1 successful hit of the global optimum. However, this is still not a fair comparison by doing so, since P3 accumulates the NFE, while tradition EA framework does not when determining the population size. To sum up, the requirement of 10 consecutive successful hits favors P3, while merely 1 successful hit favors traditional EAs. Therefore, we cannot figure out a better way to compare DSMGA-II with P3 but listing results under both 10- and 1-hit requirements. Here we only show the results for the largest problems in Table 1.
\begin{table}[!h]
\begin{center}
\begin{tabular}{|c|c|c|c|}
  \hline
   \multirow{2}{*}{\textbf{Problems}} & \multicolumn{2}{|c|}{\textbf{DSMGA-II}} & \multirow{2}{*}{\textbf{P3}}\\ \cline{2-3}    & \textbf{10-hit} & \textbf{1-hit} & \\ \hline\hline
  \textbf{Concatenated trap, $\ell=400$} & \multicolumn{1}{r|}{54} & \multicolumn{1}{r|}{36} & \multicolumn{1}{r|}{71} \\ \hline 
 \textbf{Cyclic trap, $\ell=400$} & \multicolumn{1}{r|}{125} & \multicolumn{1}{r|}{66} & \multicolumn{1}{r|}{143}\\ \hline
  \textbf{Folded trap,} $\ell=240$ & \multicolumn{1}{r|}{134} & \multicolumn{1}{r|}{117} & \multicolumn{1}{r|}{6831}\\ \hline
  \textbf{NK-S1,} $\ell=400$ & \multicolumn{1}{r|}{877} & \multicolumn{1}{r|}{408} & \multicolumn{1}{r|}{1900}\\ \hline
  \textbf{NK-S3,} $\ell=400$ & \multicolumn{1}{r|}{537} & \multicolumn{1}{r|}{323} & \multicolumn{1}{r|}{2103}\\ \hline
  \textbf{NK-S5,} $\ell=400$ & \multicolumn{1}{r|}{61} & \multicolumn{1}{r|}{53} & \multicolumn{1}{r|}{400}\\ \hline
  \textbf{Ising spin-glass,} $\ell=400$ & \multicolumn{1}{r|}{223} & \multicolumn{1}{r|}{159} & \multicolumn{1}{r|}{183}\\ \hline
  \textbf{MAX-SAT,} $\ell=100$ & \multicolumn{1}{r|}{208} & \multicolumn{1}{r|}{68} & \multicolumn{1}{r|}{151}\\ \hline\hline
  \multicolumn{4}{|r|}{unit: k}\\ \hline
\end{tabular}
\end{center}
\caption{Required NFE for the largest test problems of DSMGA-II and P3.}
\end{table}

\section{CONCLUSION}
This paper proposes a new evolutionary algorithm with linkage learning, called $\mbox{DSMGA-II}$. Similar to $\mbox{DSMGA}$, it adopts pairwise linkage detection and stores the information in a DSM. The linkage information is then used to construct a newly proposed linkage model, the incremental linkage set (ILS), which is expressive for problems with both overlapping and non-overlapping structures. Inspired by OM, the restricted mixing and the back mixing are designed to balance between exploration and exploitation. For each receiver, the restricted mixing chooses specific donors according to ILS constructed from the global information instead of random. The back mixing further refines the DSM using promising subsolutions and hence reduces unnecessary function evaluations. Empirical results shows that $\mbox{DSMGA-II}$ outperforms many black-box optimization algorithms such as LT-GOMEA, hBOA and P3 in terms of the number of function evaluations without compromising the scalability on several benchmark problems, including concatenated trap, folded trap, cyclic trap with overlapping, the NK-landscape problems with various degrees of overlapping, Ising spin-glass and MAX-SAT.
 
As for future work, we would like to test $\mbox{DSMGA-II}$ on problems with hierarchical structures. Also, one may think of several possible improvements, such as modifying the starting position of ILS or the receiver choosing in the back mixing, and we would like to investigate them. Finally, it is important to analyze $\mbox{DSMGA-II}$ from the theoretical perspective to fully understand its strength and weakness.\par
Our C$++$ implementation for DSMGA-II is available at\linebreak
https://teilab.ee.ntu.edu.tw/\#/resources.

\section{ACKNOWLEDGMENT}
The authors would like to thank the support by Ministry of Science and Technology in Taiwan under Grant No. MOST 103-2221-E-002-177-MY2-1.

\bibliographystyle{abbrv}
\bibliography{dsmga2}

\end{document}